\pdfoutput=1
\documentclass{article}

\usepackage[preprint]{neurips_2024}

\usepackage[utf8]{inputenc} 
\usepackage[T1]{fontenc}    
\usepackage{hyperref}       
\usepackage{url}            
\usepackage{booktabs}       
\usepackage{amsfonts}       
\usepackage{nicefrac}       
\usepackage{microtype}      
\usepackage{xcolor}         
\usepackage{enumitem}
\usepackage{xspace}
\usepackage{amsmath}
\usepackage[pdftex]{graphicx}
\usepackage{mathtools}
\usepackage{subcaption}
\usepackage{booktabs}
\usepackage{array}
\usepackage{wrapfig}
\usepackage{multirow}

\title{A Pure Transformer Pretraining Framework on Text-attributed Graphs}

\author{Yu Song, Haitao Mao, Jiachen Xiao, Jingzhe Liu, Zhikai Chen, \\
 \textbf{Wei Jin}, \textbf{Carl Yang}, \textbf{Jiliang Tang}, \textbf{Hui Liu}\\
 \{songyu5, haitaoma, xiaojiac, liujin33, chenzh85, tangjili, liuhui7\}@msu.edu \\
\{wei.jin, j.carlyang\}@emory.edu \\
}

\DeclareMathOperator*{\argmax}{arg\,max}

\begin{document}

\newcommand{\jt}[1]{\textbf{\textcolor{red}{[JT: #1]}}}
\newcommand{\carl}[1]{\textbf{\textcolor{brown}{[Carl: #1]}}}
\newcommand{\wei}[1]{\textbf{\textcolor{orange}{[Wei: #1]}}}
\newcommand{\yu}[1]{\textbf{\textcolor{green}{[YU: #1]}}}
\newcommand{\ljz}[1]{\textbf{\textcolor{blue}{[ljz: #1]}}}

\maketitle
\begin{abstract}
    Pretraining plays a pivotal role in acquiring generalized knowledge from large-scale data, achieving remarkable successes as evidenced by large models in CV and NLP. However, progress in the graph domain remains limited due to fundamental challenges such as \textit{feature heterogeneity} and \textit{structural heterogeneity}.
Recently, increasing efforts have been made to enhance node feature quality with Large Language Models (LLMs) on text-attributed graphs (TAGs), demonstrating superiority to traditional bag-of-words or word2vec techniques. These high-quality node features reduce the previously critical role of graph structure, resulting in a modest performance gap between Graph Neural Networks (GNNs) and structure-agnostic Multi-Layer Perceptrons (MLPs). Motivated by this, we introduce a feature-centric pretraining perspective by treating graph structure as a prior and leveraging the rich, unified feature space to learn refined interaction patterns that generalizes across graphs. Our framework, \textbf{G}raph \textbf{S}equence \textbf{P}retraining with \textbf{T}ransformer (GSPT), samples node contexts through random walks and employs masked feature reconstruction to capture pairwise proximity in the LLM-unified feature space using a standard Transformer. By utilizing unified text representations rather than varying structures, our framework achieves significantly better transferability among graphs within the same domain. GSPT can be easily adapted to both node classification and link prediction, demonstrating promising empirical success on various datasets.

\end{abstract}

\section{Introduction}
Transfer learning has witnessed remarkable success in recent years, particularly exemplified by the advancements in foundation models for Natural Language Processing (NLP) ~\cite{gpt4, touvron2023llama} and Computer Vision (CV)~\cite{kirillov2023segment, CLIP}. These methods typically leverage self-supervised pretraining on large-scale datasets to acquire broad, generalized knowledge, which is subsequently adapted to specific tasks and datasets through fine-tuning or in-context learning. However, the graph domain predominantly adheres to a 'one-model, one-dataset' approach, where models are tailored specifically to individual datasets and tasks, deviating from the prevailing trend of 'one model serves all'.

The pursuit of cross-dataset transfer learning on graphs faces unique challenges. The immediate one is \textit{feature heterogeneity}, i.e., the inherent mismatch in feature spaces among different datasets, as graphs denoting different data types often have features with varying dimensions and semantic meanings~\cite{oneforall, mao2024graph}. Previous methods circumvent this by ignoring the features and only transferring knowledge from the structural side~\cite{qiu2020gcc, davies2023its}, or constraining the applications in vertical domains where node/edge features are naturally aligned~\cite{Hu2019StrategiesFP, Xia2023MoleBERTRP}. Such approaches, while somewhat effective, suffer from performance loss or restricted applicability~\cite{Liu2021GraphSL}. Another challenge is \textit{structural heterogeneity}, which arises from the vastly different structural patterns across various graphs, leading to out-of-distribution scenarios and potential negative transfer~\cite{Wang2024SubgraphPT}. A typical example is the varying degrees of homophily across graphs~\cite{mao2024demystifying}. Together, these challenges pose substantial obstacles to the development of a flexible, generalizable model capable of effective pretraining and knowledge transfer across diverse downstream datasets.

Recent efforts ~\cite{Chen2023ExploringTP, oneforall, tan2024walklm, Chien2021NodeFE, huang2023prodigy} have been made to tackle feature heterogeneity by leveraging large language models (LLMs) to unify the feature spaces of text-attributed graphs (TAGs). They replace traditional shallow features like word2vec and tf-idf with language model-enhanced features and have demonstrated impressive empirical success, represented by the improved performance on graph-related tasks and the reduced gap between purely feature-based approaches (e.g., an MLP) and graph-tailored models (e.g., Graph Neural Networks)~\cite{Chen2023ExploringTP}. This trend motivates us to consider a conceptual shift from an era predominantly focused on graph structure to a  feature-centric paradigm, suggesting the potential to enhance knowledge transfer by effectively leveraging LLM-based unified features.

Following this perspective, the key principle is to identify a fundamental unit that effectively encodes graph information and generalizes across the LLM-unified feature space. In this study, we propose to learn a pairwise function that models the proximity of node pairs, thereby capturing the interactions embedded in the graph structure. Inspired by works in CV, NLP and the graph domain~\cite{He2021MaskedAA, Devlin2019BERTPO, tang2021graph}, we adopt a reconstruction-based objective as the pretext task due to its efficacy in enforcing a robust understanding of dependencies within various data schemes. Specifically, given a (masked) center node and its context, we aim to reconstruct the feature of the center node using its context. 
This approach presents two key challenges. The first challenge is \textit{how to construct the appropriate context for graphs}? Unlike sequential data such as language or grid-based data like images, graphs represent non-Euclidean structures, making it non-trivial to create a context for reconstruction. The second challenge is \textit{how to reconstruct the center node based on its context}? Specifically, it involves identifying an appropriate architecture that can accurately interpret the context nodes and a loss function that effectively guides the learning process of the model. 

To address the first challenge, we propose using node sequences generated by random walks as contexts, where the entire sequence forms the receptive field, and the order of nodes preserves proximity information~\cite{Grover2016node2vecSF, Perozzi2014DeepWalkOL}. By doing so, the original graph topology serves as a prior to retrieve relevant context nodes, facilitating effective proximity modeling for accurate reconstruction of masked features.
For the second challenge, we employ a standard Transformer architecture due to its flexibility in modeling sequences with self-attention and its proven transferability across domains like CV and NLP~\cite{kirillov2023segment, Devlin2019BERTPO}. To handle the multi-dimensional and continuous nature of LM-produced features, we adopt a cosine similarity-based objective to measure reconstruction error instead of cross-entropy~\cite{hou2022graphmae}.

Putting it all together, we propose \textbf{G}raph \textbf{S}equence \textbf{P}retraining with \textbf{T}ransformer (GSPT), where a standard Transformer is used alongside a feature reconstruction objective to learn a unified model for node representations. 
Our framework highlights the use of text-based representations rather than the diverse structures of different graphs, leading to enhanced transferability and reduced risk of negative transfer caused by structural shifts. 
To evaluate the effectiveness of our method, we perform self-supervised training on the largest graph available ogbn-papers100M~\cite{hu2020open}, and apply the pretrained framework on various downstream datasets. Experimental results reveal that our GSPT excels in in-context node classification and link prediction, showcasing effective knowledge transfer from pretraining data to downstream tasks. Furthermore, we observe a notable trend of improvement with increased pretraining data, highlighting the potential of the proposed framework.
Our findings deepen the understanding of the LLM-unified feature space in graph data and provide valuable insights into the development of a versatile and generalizable graph foundation model.

\section{Preliminary Study}
\begin{figure}[ht]
    \centering
        \begin{subfigure}[b]{0.35\textwidth}
        \centering
        \includegraphics[width=\textwidth]{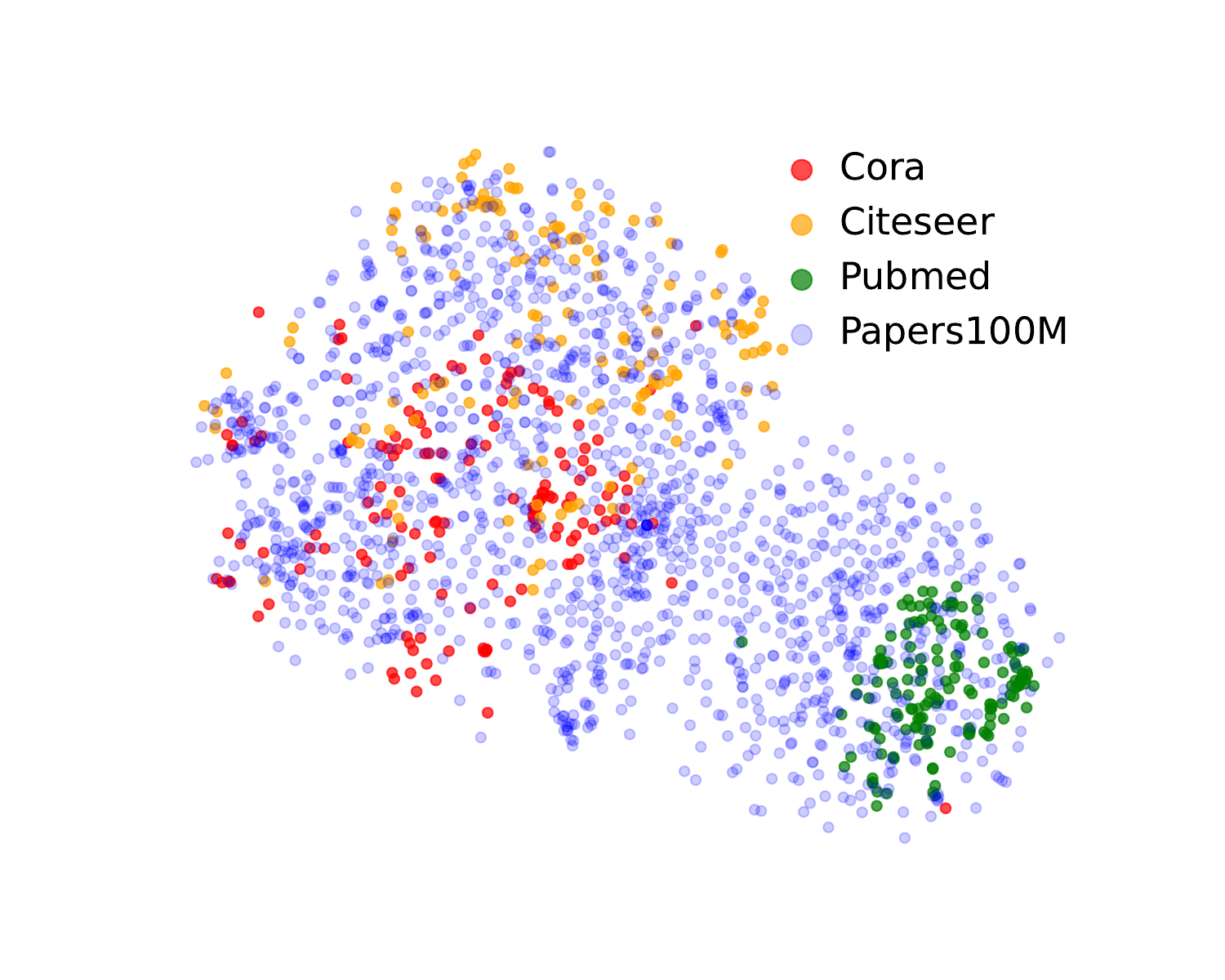}
        \caption{t-SNE visualization of feature space}
        \label{fig:figure2}
    \end{subfigure}
        \hfill
    \begin{subfigure}[b]{0.3\textwidth}
        \centering
        \includegraphics[width=\textwidth]{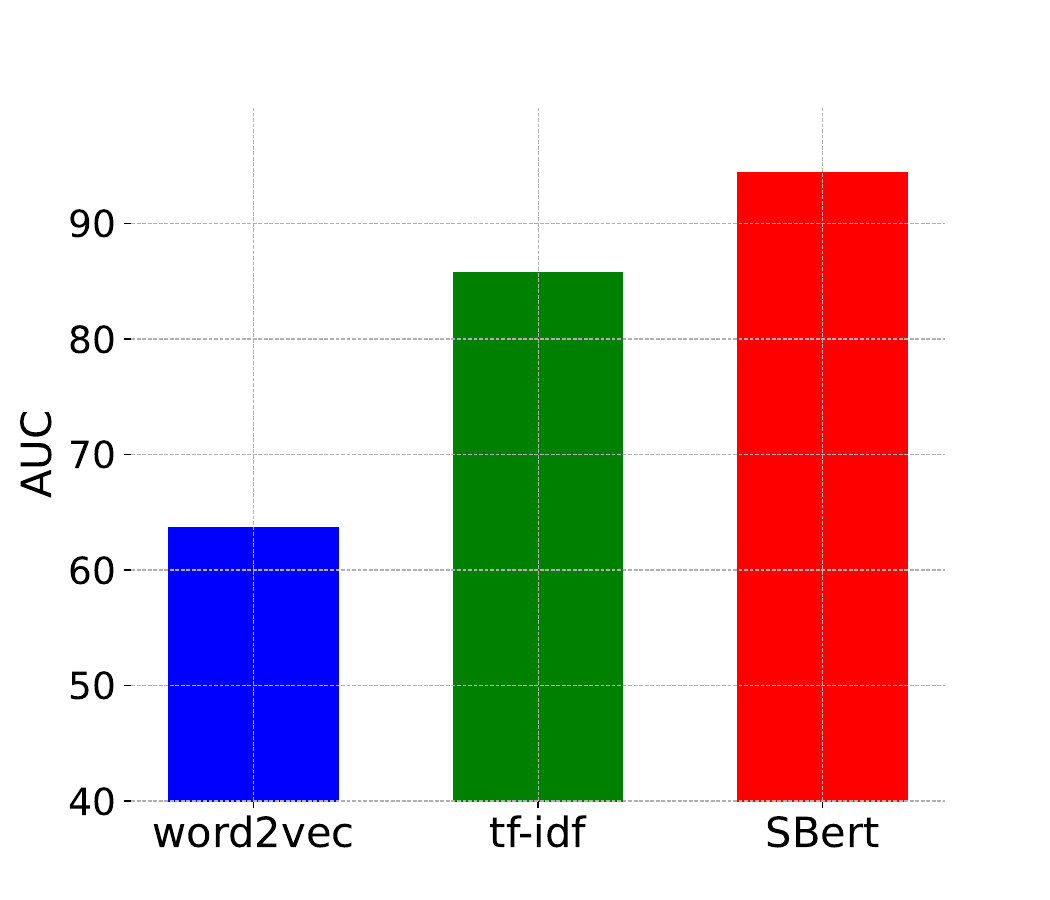}
        \caption{Link prediction}
    \end{subfigure}
    \hfill
    \begin{subfigure}[b]{0.3\textwidth}
        \centering
        \includegraphics[width=\textwidth]{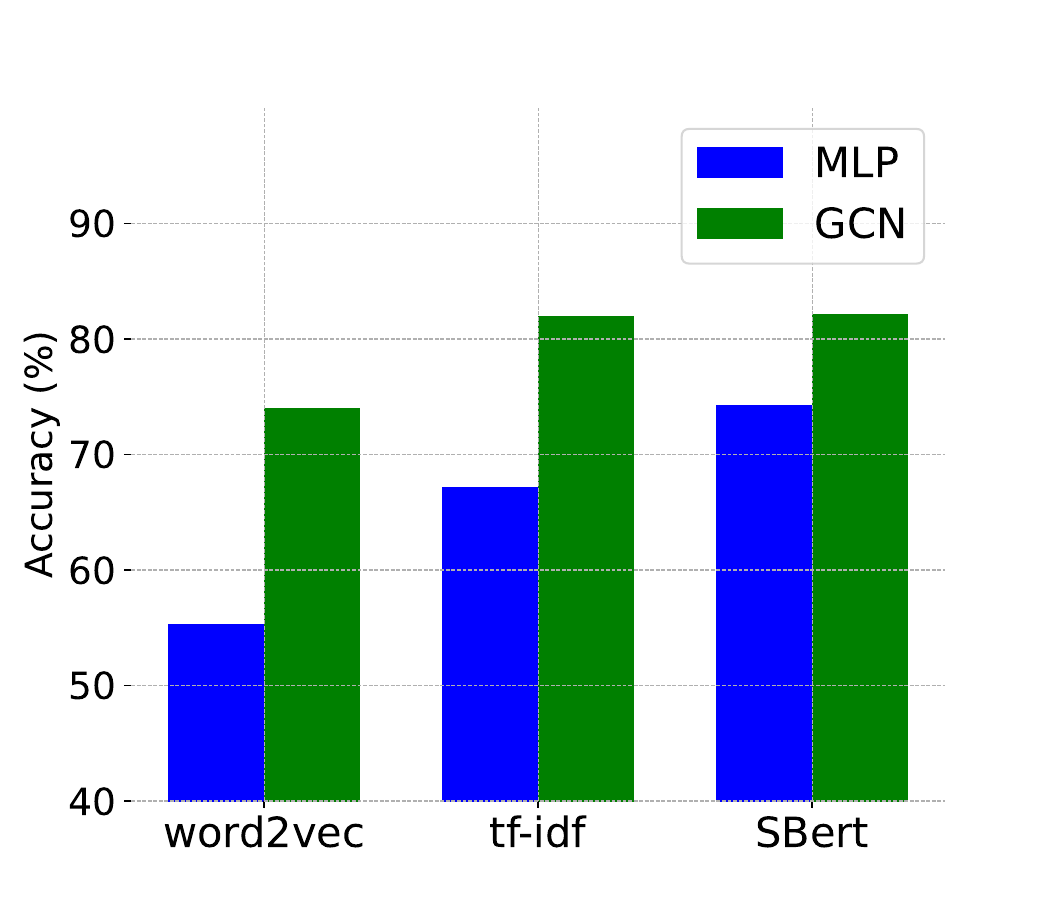}
        \caption{Node classification}
        \label{fig:figure1}
    \end{subfigure}

    \caption{(a) SentenceBert provides a unified feature space for different datasets under the same domain. The node features of three small citation graphs, i.e., Cora, Citeseer, and Pubmed, can be well covered by ogbn-papers100M, a large-scale citation network containing papers from a vast variety of research topics.
    (b) Advanced text embeddings are better at predicting the missing edges compared with shallow features. (c) Advanced embeddings reduce the performance gap on node classification between GCN and MLP. Experiments are conducted on Cora.}
    \label{fig:preliminary}
\end{figure}

In this section, we aim to answer the question: \textit{What can be better transferred across graphs?} In general, a graph \(G=(A, X)\) is comprised of two ``modalities", i.e., the structure space represented by the adjacency matrix \(A\), and the feature space of \(X\). 
Previous studies~\cite{mao2024demystifying, mao2024revisiting} show that the structure space can hardly be unified since the connections may be formed due to vastly different principles even for graphs in similar domains. For instance, in a friendship network, nodes tend to form edges with others of the same gender, whereas the pattern is reversed in a dating network. Such inherent shift poses significant challenges to the transfer learning on the structure space, even leading to negative transfer~\cite{Wang2024SubgraphPT,  lin2022spectral}. On the other hand, the feature space \(X\) can be effectively unified with a powerful LM, as demonstrated in Figure ~\ref{fig:preliminary} (a), where a pretrained SentenceBERT~\cite{reimers2019sentence} is used to generate node embeddings for different datasets within the citation domain.  

Therefore, we ask: can we focus on transferring knowledge from the feature space, while reconstructing the structure information based on the unified features? To answer this question, we conduct two sets of experiments to determine the extent to which structure can be reconstructed from the feature information. Based on Figure ~\ref{fig:preliminary}, the SentenceBert embeddings significantly facilitate the structure reconstruction, demonstrated by the improved link prediction performance (b) and the reduced gap between GCN and MLP in node classification (c). 

The aforementioned observations motivate us to design our pretraining framework from a feature-centric perspective: we consider graph structure as a prior and utilize features $X$ to capture the  fine-grained topology that generalizes across graphs. In particular, we propose to model the graph structure via pairwise relationships based on the unified feature space derived from LLM embeddings and transfer this function to datasets within similar domains. The pairwise relationship is useful because it is closely related to various graph-based tasks, e.g., for node classification, we can frame the problem as "whether two nodes belong to the same class"; for link prediction, the question can be framed as "whether there exists an edge between a pair of nodes". Compared with strictly adhering to the fixed inductive bias of graph topology, this approach inherently avoids the potential negative transfer due to the structure mismatch and has the potential to improve with advances in LLMs to unify the node attributes.

\section{Method}

\subsection{An Overview}
\begin{figure}[ht!] 
    \centering
    \includegraphics[width=1.0\textwidth]{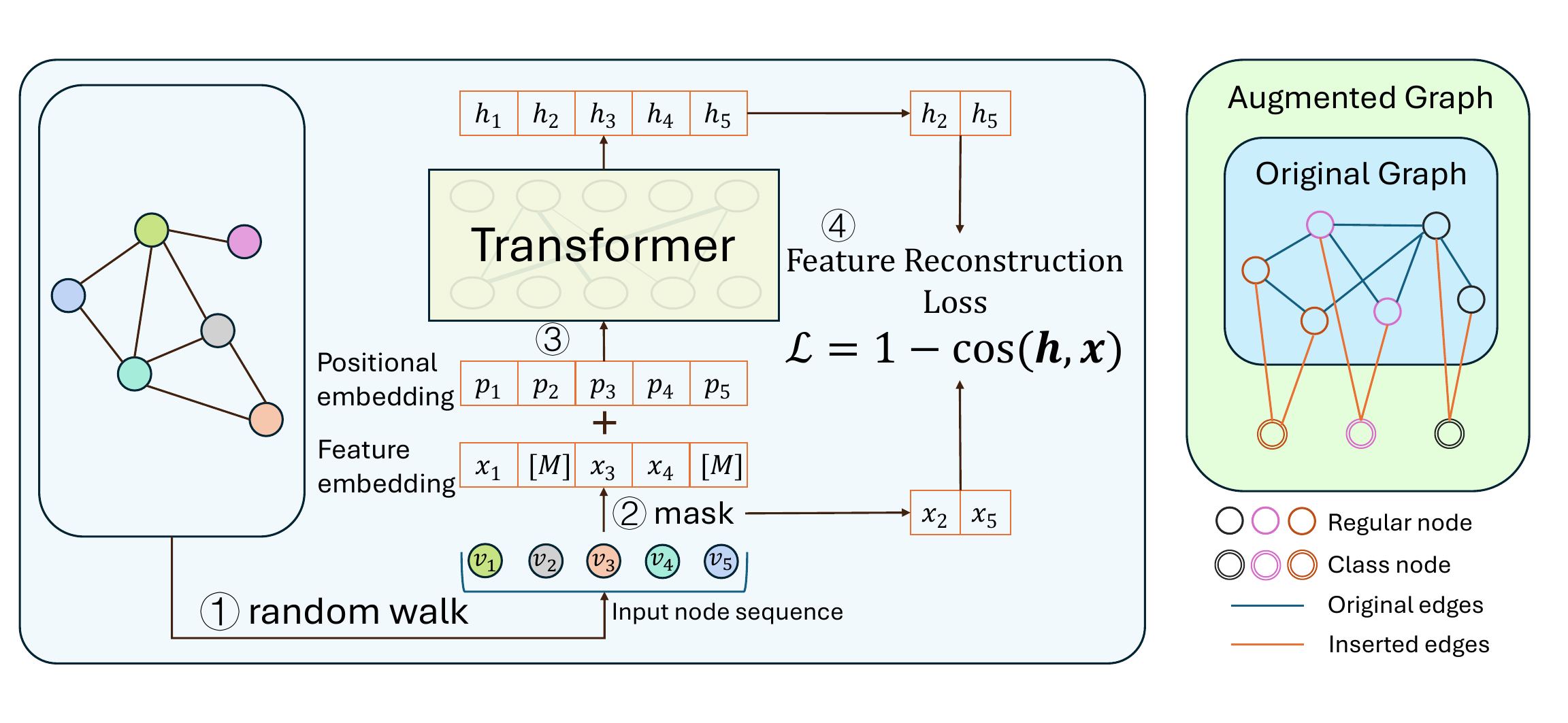} 
   
    \caption{The overall framework of Graph Sequence Pretraining with Transformer (GSPT). Left: the pretraining consists of four steps: (1) generate node sequences from the graph using random walk; (2) randomly replace a portion of node features with [MASK]; (3) feed the input sequence into the Transformer and (4) compute the feature reconstruction loss with cosine similarity. Right: We construct the augmented graph by adding class nodes to the original graph and connecting correponding node pairs. GSPT performs in-context node classification by comparing the cosine similarity between the representations of regular nodes and class nodes. }
     \label{fig:network_structure}
\end{figure}

In this section, we introduce our feature-centric pretraining framework Graph Sequence Pretraining with Transformer (GSPT), a direct implication of the preliminary study. 
It consists of two major components. First, we generate \textit{node contexts} from the graph through random walks. This context encapsulates the structural information of the graph and is tailored to the specific task. Next, we feed the context into a standard Transformer and perform masked feature reconstruction on the large-scale pretraining dataset. This approach enables the Transformer to effectively model pairwise relationships within a unified feature space, facilitating seamless transfer to downstream datasets. Next we will illustrate our framework using the node classification task and then extend it to link prediction in Appendix~\ref{sec:extension-link-prediction}. The overall framework of GSPT is illustrated in Figure ~\ref{fig:network_structure}.

\subsection{Context Construction}
\label{sec:context-construction}

Following the preliminary study, our objective is to model the pairwise relationships between nodes and transfer this knowledge to unseen datasets. Given the typically large number of nodes in a graph, it is impractical to model relationships between all possible node pairs. Consequently, the standard approach is to define a context and focus only on the nodes within this context for modeling purposes. For example, Graph Neural Networks (GNNs) employ a K-hop neighborhood, where only nodes within a K-hop radius are utilized for message passing, and nodes outside this range have no influence on the central node. However, such fixed neighborhoods may not be optimal due to noisy connections and the potential for correlated nodes to fall outside the defined context, especially when the neighborhood size is small (e.g., K=2). Moreover, extracting the ego network poses a significant computational burden, particularly for large-scale graphs with high average degrees.

Alternatively, we use a random walk approach to construct node contexts. From the effectiveness perspective, this removes the fixed graph inductive bias and generates a connected component instead, which can be thought of as recalling the coarsely relevant nodes from the original graph. Additionally, the order of nodes in the sequence provides fine-grained information about their proximity. In terms of efficiency, random walks can be easily parallelized, making them capable of handling large graphs. Specifically, we adopt the random walk approach used by Node2vec~\cite{Grover2016node2vecSF}, where two parameters, \(p\) and \(q\), are used to adjust the properties of the generated node sequences, allowing for bias towards either a local or global view of the graph. During each training epoch, we generate a random walk sequence starting from each node in the graph to ensure that all nodes are included for pretraining.

\subsection{Transformer as Backbone}
\label{sec:backbone}
With random walks as the contexts, our next step is to identify a suitable backbone to effectively learn the interaction patterns within each context. As discussed earlier, GNNs may not be the best fit for this purpose due to their fixed graph inductive biases. While Graph Transformers aim to address these issues by integrating self-attention with fixed graph structures, they often become complex and are typically tailored for specific tasks such as graph classification~\cite{Rampek2022RecipeFA}.

In contrast, the Vanilla Transformer~\cite{Vaswani2017AttentionIA} stands out as an ideal architecture for our task due to several key advantages. Firstly, Transformers enable flexible relational modeling through self-attention, allowing for dynamic and adaptable feature propagation. Secondly, their straightforward design simplifies implementation and scaling compared to more intricate variants. Lastly, they integrate several successful components in deep learning such as layer normalization and residual connections, greatly reducing the design space of the framework.

\subsection{Masked Feature Reconstruction}
\label{sec:masked-feature-reconstruction}

Inspired by the huge success of self-supervised learning (SSL) methods in CV and NLP that have proven to improve performance with more data, we aim to identify an appropriate pretext task for graphs. Given a unified feature space via language models, feature reconstruction-based SSL has the potential to continually incorporate more data for pretraining as it becomes available. Specifically, with sequences as inputs and  Transformer as the backbone, the most popular SSL methods are Masked Language Modeling (MLM) in BERT~\cite{Devlin2019BERTPO} and Next Token Prediction (NTP) in GPT~\cite{Brown2020LanguageMA}.
We choose the BERT approach here due to its bidirectional attention, which is favored in the random walk setting where context from both preceding and succeeding nodes enhances prediction.

Specifically, given a random walk of length \(l\): \( rw = [v_0, v_1, ..., v_{l-1}] \), the input sequence is constructed by extracting the corresponding node features \(x_i\in \mathbb{R}^{d}\), added by the positional embeddings \(p_i\in \mathbb{R}^{d}\), i.e., \(h^0 = [x_0 + p_0, x_1 + p_1, ..., x_{l-1} + p_{l-1}]\). Before feeding to the Transformer, we randomly replace a ratio of nodes in the sequence with a [MASK] token. The goal is to reconstruct the raw node feature of the masked nodes based on the output of the Transformer.
Unlike the original MLM task in BERT, where a cross-entropy loss is used to predict the ID of the masked tokens, we use cosine similarity to measure how well the feature is reconstructed based on the context.
Formally, the node representations for a given input batch \( H^0 \in \mathbb{R}^{b \times l \times d} \) of \( b \) nodes are obtained through the following steps:
 \begin{equation}
    H^L = \text{TRM}(H^0)
 \end{equation}
  \begin{equation}
    H_{node} = \text{Pooling}(H^L)
 \end{equation}
  where \(\text{TRM}\) is a Transformer with $L$ layers, and \(H^L\) is the output of the final layer. \(\text{Pooling}\) denotes a pooling function that reduces the output \(H^L \in \mathbb{R}^{b\times l \times d}\) to node representations \(H_{node}\in \mathbb{R}^{b\times d}\). For simplicity, we use \(h_{i}\) to denote the i-th row of pooled representations \(H_{node}\). In our implementation, the node representation \( h_i \) for node \( v_i \) is computed by taking the average of all embeddings of \( v_i \) in \( H^L \) to enhance the representation quality.

Finally, given a masked sequence of nodes, the loss function for the feature reconstruction task  is computed as follows:
\begin{equation}
    \mathcal{L}_{\text{node}} = \frac{1}{m} \sum_{i=1}^{m} \left(1 - \cos(h_i, x_i)\right)
\end{equation}
where \( \cos(h_i, x_i) \) denotes the cosine similarity between the reconstructed representation \( h_i \) and the original feature \( x_i \), and \( m \) is the number of masked nodes in the sequence.

\subsection{Negative sampling}
\label{sec:negative-sampling}

A random walk sequence forms a local connected component and contains nodes that are largely correlated. Due to the homophily assumption, nodes in proximal positions are likely to provide useful information for the reconstruction of masked nodes, creating a shortcut for the pretext task. Consequently, the attention mechanism may not be fully activated to select the most relevant neighbors from the context to attend to, as even a simple mean pooling could yield a decent reconstruction error in a homophilic neighborhood~\cite{hou2022graphmae}.

To tackle this issue, we introduce noisy nodes into the input sequence as "distractor nodes." This ensures that the model cannot achieve good reconstruction without learning to attend to only the truly relevant nodes. In practice, we randomly select a node from the graph and repeat it \( K \) times at the end of each input sequence, where \( K \) is a random number from \([0, l]\). 
Formally, the random walk sequence with distractor nodes is given by:
\begin{equation}
    rw = [v_0, v_1, ..., v_{l-K-1}, v_d, v_d, ..., v_d] 
\end{equation}
where \(v_d\) is the distractor node.
In the experiments, we show that the distractor nodes play an important role in the pretraining of the model.

\subsection{Enabling In-context Learning}

After pretraining, our framework is able to produce high-quality node representations for input graphs from similar domains. To fully utilize the knowledge of the pretrained model, we design an in-context learning framework to address few-shot learning tasks using support examples, inspired by the graph prompts in ~\cite{oneforall, huang2023prodigy}.

This in-context learning process is illustrated in the right part of Figure ~\ref{fig:network_structure}. For an N-way K-shot task, we first introduce N class nodes to the original graph \(G_{\text{ori}}\). We then connect the K-shot examples to their corresponding class nodes, creating an augmented graph \(G_{\text{aug}}\). Similar to the pretraining process, we transform the augmented graph into sequences using random walks. Our pretrained backbone then performs feature propagation on these node sequences, merging the features of regular nodes and class nodes through the self-attention mechanism.
The output node embeddings, both for regular and class nodes, are used for prediction by comparing the cosine similarity between the test nodes and all class nodes. 
Given the output embeddings \( \{ s_1, s_2, \ldots, s_N \} \) for the $N$ class nodes and the embedding \( t \) for a test node, the predicted class is:
\begin{equation}
\hat{y} = \argmax_{i \in \{1, \ldots, N\}} \cos(t, s_i)
\end{equation}
Through this approach, our framework is able to make predictions on any unseen test graph with novel labels, without modifying the parameters of the pretrained model.

\section{Experiment}

In this section, we conduct experiments to validate the effectiveness of our proposed GSPT framework. 
Through the experiments, we aim to answer the following research questions: 
\textbf{RQ1:} Can our proposed pretraining framework achieve cross-graph knowledge transfer?
\textbf{RQ2:} What is the underlying mechanism that promotes the positive transfer of GSPT? 
\textbf{RQ3:} How do different negative sampling strategies affect performance? 
\textbf{RQ4:} Can our framework benefit from increasing the data scale used for pretraining?

\subsection{Datasets}
\label{sec:datasets}

We evaluate our framework using a variety of citation datasets. For pretraining, we utilize the ognb-papers100M~\cite{hu2020openOGB} dataset, which comprises over 100 million nodes and 1.6 billion edges, making it the largest publicly available graph dataset. To address efficiency issues, we use the METIS algorithm to partition the entire graph into approximately 10,000 smaller graphs during pre-processing, each containing around 10,000 nodes. During pretraining, we randomly select one partition to construct a batch.
For downstream tasks, we assess the performance on both node classification and link prediction using four well-known citation graphs: Cora, Citeseer, Pubmed, and Arxiv23. We neglect ogbn-arxiv in our evaluation as it is covered by ogbn-papers100M. To ensure a consistent and unified feature space among all datasets, we generate text embeddings for the raw text associated with each node using SentenceBERT~\cite{reimers2019sentence}.

\subsection{Few-shot Node Classification}
\subsubsection{Experimental Setup}

\textbf{Setup}. We begin by evaluating our pretraining framework on the node classification task. In line with ~\cite{oneforall}, we adopt a few-shot in-context learning setting to create N-way K-shot tasks. Specifically, we randomly select \(N\) classes from the dataset's class set and generate a K-shot prompt, where \(K\) labeled nodes are randomly chosen for each of the \(N\) classes. All K-shot examples are drawn from the original training split of the graphs. The validation and test sets are constructed by filtering the nodes of the selected classes from the original validation and test sets, respectively. In all experiments, we generate 100 templates for each of the N-way K-shot tasks.

\textbf{Baselines}. We compare our method against four groups of baselines. The first group includes end-to-end methods, represented by GCN~\cite{kipf2017semisupervised} and GAT~\cite{Velickovic2017GraphAN} that are directly trained on the downstream datasets. The second group comprises feature-based approaches, which perform node classification by comparing the cosine similarity between the test nodes and prototype vectors denoting the centroid of the K-shot prompt nodes. To make the features contextualized w.r.t. the graph structure, we adopt the degree-normalized Laplacian to perform feature propagation following ~\cite{Wu2019SimplifyingGC}. The third group includes transfer learning results of two representative graph self-supervised learning methods, DGI~\cite{Velickovic2018DeepGI} and GraphMAE~\cite{hou2022graphmae}. Both methods are pretrained on ogbn-papers100M~\cite{hu2020openOGB}, and are evaluated via cosine similarity using their generated embeddings.
Lastly, we compare our GSPT with in-context learning baselines, including Prodigy~\cite{huang2023prodigy} and OneForAll~\cite{oneforall}. Prodigy utilizes neighbor matching and supervised training on the MAG240M dataset, while OneForAll is trained on ogbn-arxiv with few-shot learning templates directly mimicking the downstream task.

\textbf{Our method}. 
Depending on how to construct the features for the class nodes, we have two variants of our proposed framework GSPT, namely, GSPT-void and GSPT-desc. GSPT-void means initializing the features for all class nodes with zero vectors. In contrast, in GSPT-desc, we use GPT4~\cite{gpt4} to generate a text description for each class of the downstream dataset, and use the SentenceBert~\cite{reimers2019sentence} embeddings of the descriptions to initialize the class nodes, following ~\cite{oneforall}.  

\subsubsection{Result Comparison}
 Table ~\ref{tab:node_performance_comparison} shows the performance comparison between the four groups of methods on N-way-3-shot tasks with varying Ns. We make the following observations:
 
 \textbf{GSPT exhibits strong in-context learning ability}. 
 Both variants of GSPT surpass the two ICL baselines by a significant margin, indicating the effective knowledge transfer of our proposed framework.
 Among the two variants, GSPT-desc consistently outperforms GSPT-void, suggesting that our pretrained model can further leverage the additional information provided in the class descriptions. Surprisingly,  GSPT-desc achieves the highest accuracy in most scenarios, even outperforming expert GNN models directly trained on the downstream dataset. This indicates that when label information is sparse, in-context learning may be a better solution than training specific models.

 \textbf{Feature-based approaches serve as a decent baseline.} Feature-based methods offer a straightforward implementation of model-free classification and can be practical in low-resource scenarios. Among these methods, SentenceBERT outperforms other shallow embeddings, significantly reducing the gap to end-to-end models. This indicates that using advanced language model embeddings as node features, combined with simple message-passing, allows a model-free method to generate highly contextualized and discriminative representations for node classification. 

\textbf{Existing self-supervised learning methods lead to negative transfer.} We assess the effectiveness of self-supervised learning methods in transfer learning scenarios. Ideally, with graph-aware pretraining on a large-scale dataset, the graph encoder should produce higher-quality node representations than those generated by simple feature-based approaches. However, our observations indicate that methods like DGI and GraphMAE consistently underperform when compared to the SentenceBert baseline, showcasing negative transfer. We attribute this to the structural shift from the pretraining graph to the downstream datasets, particularly highlighted by the poor performance on Arxiv23, which exhibits different properties compared to the pretraining dataset Papers100M (See Appendix ~\ref{sec:dataset} for details). Conversely, our GSPT consistently enhances the SentenceBert baseline, demonstrating its capability to promote positive transfer and effectively leverage the pretrained knowledge for improved performance.

Overall, GSPT demonstrates strong cross-graph transferability, achieving better or comparable accuracy compared to end-to-end methods on few-shot node classification tasks across different datasets, without modifying any pretrained parameters (Answer to \textbf{RQ1}).

\subsubsection{Analysis of GSPT's in-context capability} 
We aim to understand the strong in-context learning ability of GSPT from the perspective of attention mechanisms, using GSPT-desc as an example. Specifically, using each class prototype as the query, we visualize the attention distribution computed by the Transformer on its context nodes of different classes. As illustrated in Figure ~\ref{fig: pre_part_conf}, the pretrained Transformer learns to adaptively propagate messages within the augmented graph, i.e., enabling class nodes to selectively attend to regular nodes that share the same ground-truth labels, and vice versa (Answer to \textbf{RQ2}). Consequently, the similarity between intra-class nodes increases, while it decreases for inter-class nodes. In contrast, the Transformer without pretraining relies solely on the graph structure for message passing, and thus the performance is bounded by the homophily of the original graph.

\begin{table}[ht]
    \centering
    \caption{Performance comparison of few-shot node classification on citation datasets. We report the Accuracy (\%) on 100 sampled tasks with 3-shot prompts.}
    \resizebox{\linewidth}{!}{
    \begin{tabular}{llllllllll}
        \toprule
        Methods & \multicolumn{3}{c}{Cora} & \multicolumn{2}{c}{Citeseer} & \multicolumn{1}{c}{Pubmed} & \multicolumn{3}{c}{Arxiv23}\\
        \cmidrule(lr){2-4} \cmidrule(lr){5-6} \cmidrule(lr){7-7} \cmidrule(lr){8-10}
        & 2-way & 5-way & 7-way & 3-way & 6-way & 3-way & 3-way & 5-way & 10-way \\
        \midrule
        \multicolumn{10}{c}{End-to-end Methods} \\
        \midrule
        GCN & 91.90 & 77.22 & 71.60 & 80.38 & 67.46 & 66.56 & \textbf{88.34} & 82.14 & 67.98 \\
        GAT & \textbf{93.00} & 78.19 & 72.28 & 80.96 & 68.02 & 66.82 & 87.75 & \textbf{83.80} & 69.21 \\
        \midrule
        \multicolumn{10}{c}{Feature-based Approach} \\
        \midrule
        Word2vec & 78.81 & 56.73 & 48.82 & 69.49 & 53.74 & 56.09 & 60.57 & 59.01 & 44.05 \\
        Tf-idf & 87.96 & 70.98 & 64.27 & 76.40 & 62.78 & 63.76 & 
        79.74 & 75.77 & 60.52 \\
        SentenceBert & 89.39 & 74.56 & 67.91 & 79.10 & 66.70 & 64.55 &
        83.66 & 77.01 & 62.94 \\
        \midrule
        \multicolumn{10}{c}{{Self-supervised Transfer}} \\
        \midrule
        DGI & 62.47 & 33.51 & 26.47 & 40.18 & 23.20 & 41.27 & 38.63 & 28.71 & 13.09 \\
        GraphMAE & 88.82 & 73.52 & 66.77 & 75.80 & 62.54 & 62.39 & 27.21 & 24.77 & 11.07 \\
        \midrule
        \multicolumn{10}{c}{In-context Learning} \\
        \midrule
        Prodigy & 73.57 & 62.54 & 58.31 & 69.81 & 61.63 & 60.24 & 68.35 & 58.43 & 42.14 \\
        OFA & 72.35 & 60.57 & 56.47 & 67.99 & 59.53 & 59.88 & 69.23 & 58.23 & 42.68\\
        \midrule
         GSPT-void & 91.92 & 77.00 & 71.40 & 80.02 & 68.37 & 65.51 & 86.06 & 78.26 & 63.12 \\
        GSPT-desc & 92.89 & \textbf{80.55} & \textbf{75.61} & \textbf{82.23} & \textbf{70.88} & \textbf{70.85} & 87.67 & 81.46 & \textbf{69.40}\\
        \bottomrule
    \end{tabular}}
    \label{tab:node_performance_comparison}
\end{table}

\begin{minipage}[c]{0.55\textwidth}
    \centering
    \includegraphics[width=\linewidth]{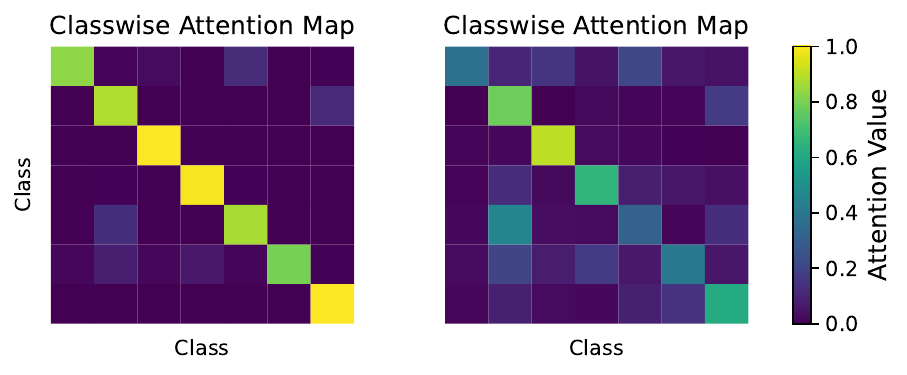}
    \captionof{figure}{Attention map on classes of Cora. Left: attention weights obtained by the pretrained Transformer. Right: attention weights w/o pretraining.}

    \label{fig: pre_part_conf}
\end{minipage}
\hfill
\begin{minipage}[c]{0.43\textwidth}
    \centering
    \captionof{table}{Performance comparison of link prediction tasks on citation datasets. We adopt MRR as the evaluation metric. }
    \vspace{5pt}
    \resizebox{\linewidth}{!}{
\begin{tabular}{ccccc}
        \hline
        Method & Cora & Citeseer & Pubmed & Arxiv23 \\
        \hline
        \multicolumn{5}{c}{\textbf{Feature-based}} \\
        \hline
        SentenceBert & 55.67 & 64.19 & 63.47 & 70.51 \\
        Node2vec & 58.06 & 46.26 & 53.53 & 56.50 \\
        \hline
        \multicolumn{5}{c}{\textbf{GNN}} \\
        \hline
        GCN & 70.25 & 64.88 & 78.96 & 79.39 \\
        GraphSAGE & 64.12 & 63.95 & 79.16 & 78.04 \\
        \hline
        \multicolumn{5}{c}{\textbf{GSPT}} \\
        \hline
        GSPT-TFS & 68.61 & 71.16 & 79.48 & 79.67 \\
        GSPT-pretrained & \textbf{72.67} & \textbf{75.16} & \textbf{81.59} & \textbf{82.13} \\
        \hline
    \end{tabular}}
    \vspace{0.1cm}
    \label{tab:link_performance_comparison}
\end{minipage}\hfill

\subsection{Link prediction}

\textbf{Setup}. For the link prediction task, we construct the dataset by randomly sampling 80\%, 10\%, and 10\% of the edges in the graph for the training, validation, and test sets, respectively. We use the Mean Reciprocal Rank (MRR) as our evaluation metric. Unlike the in-context node classification setting, we finetune the pretrained GSPT on the downstream dataset to better adapt it to the specific properties of the individual dataset.

\textbf{Baselines}. We compare our method against two categories of baselines. For feature-based approaches, we use the SentenceBert embeddings and Node2Vec embeddings which are solely based on the feature and structure, respectively. For GNNs, we include GCN and GraphSAGE for comparison. We do not compare with GNN4LP methods like ~\cite{BUDDY, wang2023neuralNCN} as they explicitly incorporate structural information when computing the edge scores, and do not reflect the capacity of node representations. Subgraph-based methods like OneForAll~\cite{oneforall} and Prodigy~\cite{huang2023prodigy} present severe efficiency issues, which makes them non-applicable to large-scale pretraining for link prediction. For our method, we evaluate both the variant trained from scratch (GSPT-TFS) and the variant fine-tuned from the pretrained checkpoint (GSPT-pretrained). In all methods, we employ an MLP on top of the generated node embeddings to compute edge scores.

\textbf{Results}.  As shown in Table ~\ref{tab:link_performance_comparison}, GSPT-pretrained surpasses all baseline methods. Notably, GSPT-pretrained consistently outperforms GSPT-TFS. This demonstrates that pretraining on a large-scale dataset allows the model to effectively learn the underlying principles of edge formation, leading to improved performance when transferred to downstream datasets (Answer to \textbf{RQ1}).

\subsection{Ablation Study}

\begin{wrapfigure}{r}{0.5\linewidth}
    \centering
    \vspace{-3em}
    \includegraphics[width=\linewidth]{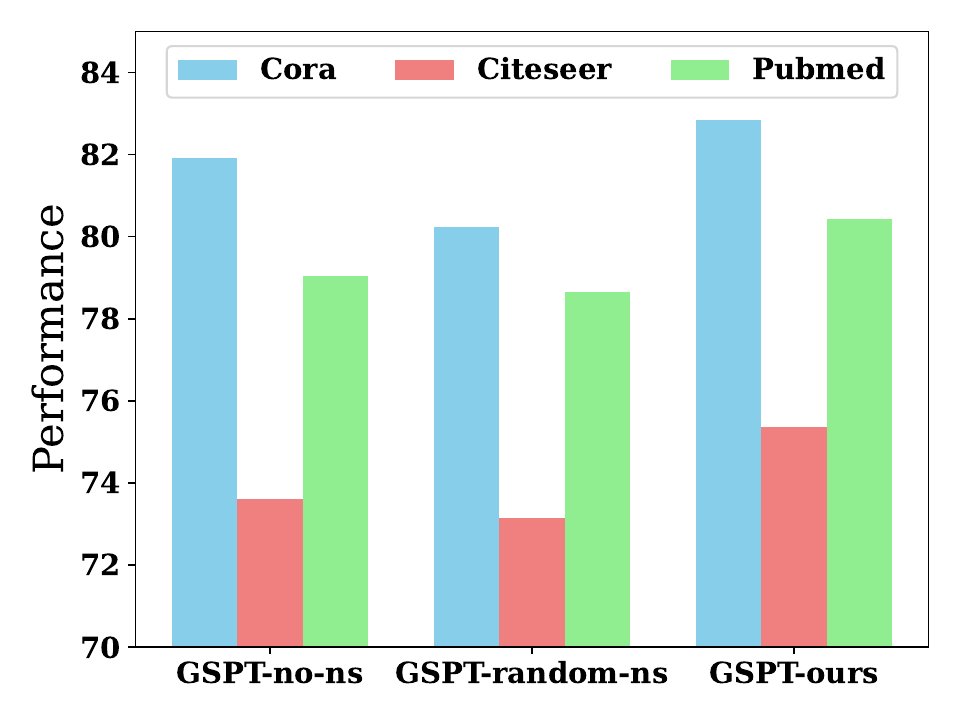}
    \setlength{\abovecaptionskip}{-0.3cm}
    \caption{Ablation studies of different negative sampling strategies.}
    \label{fig:ablation}
    \vskip -1em
\end{wrapfigure}

In this subsection, we compare three variants of the GSPT framework to demonstrate the effectiveness of our negative sampling strategy during pretraining. Specifically, GSPT-no-ns refers to pretraining the GSPT without negative sampling, i.e., using the entire random walk sequences as input. GSPT-random-ns means randomly sampling K independent nodes at the end of each sequence. In GSPT-ours, we randomly sample one negative node for each sequence, and repeat the same  node K times at the end of the sequence.
The results are presented in Figure ~\ref{fig:ablation}.
We observe that GSPT-no-ns performs worse than GSPT-ours, demonstrating that the feature reconstruction task alone cannot compel the model to effectively learn interaction patterns without the inclusion of negative samples. On the other hand, GSPT-random-ns performs even worse, as this manner introduces too much noise to the input sequences, interfering with the training. Overall, our negative sampling approach, while simple and straightforward, is an effective method to aid the pretraining process (Answer to \textbf{RQ3}).

\subsection{Scaling Effect}

\begin{figure}[h!]
    \centering
    \begin{subfigure}[b]{0.45\textwidth}
        \centering
        \includegraphics[width=\textwidth]{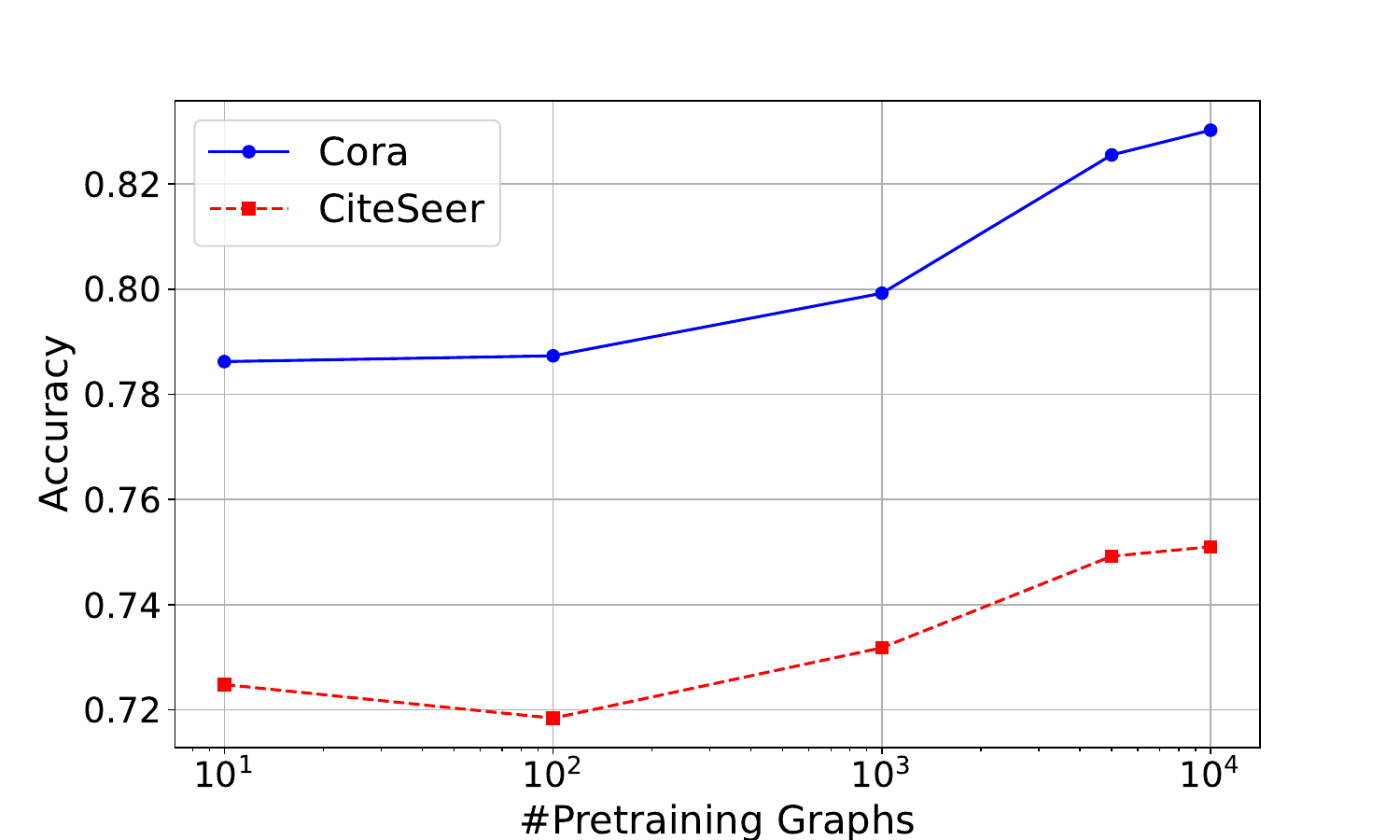} 
        \caption{Node Classification}
    \end{subfigure}
    \hfill
    \begin{subfigure}[b]{0.45\textwidth}
        \centering
        \includegraphics[width=\textwidth]{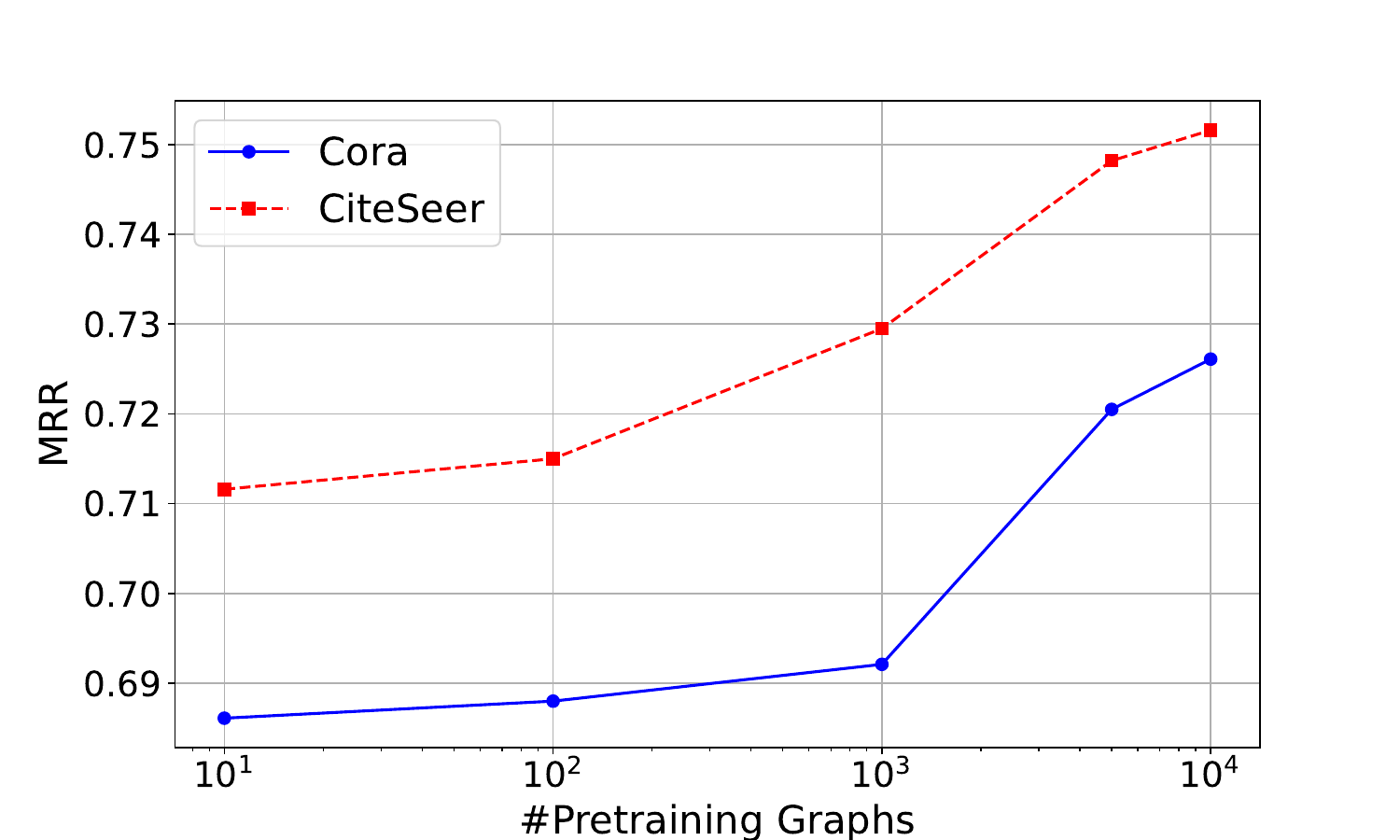} 
        \caption{Link Prediction}
    \end{subfigure}
    
    \caption{Scaling effect of GSPT. (a) Node classification performance on downstream datasets with linear probing. (b) Link prediction performance on downstream datasets via fine-tuning. X-axis denotes the number of METIS graphs used for pretraining. Empirically, GSPT improves as adding more data to pretraining.}
    \label{fig:scalilng}
\end{figure}

The NLP and CV domains have observed a scaling law in Transformer-based models, where performance on downstream tasks improves as more data is used for pretraining. To investigate whether GSPT exhibits a similar property, we conducted control experiments using different ratios of the pretraining dataset and evaluated the corresponding performance on downstream tasks. Specifically, we randomly selected varying proportions of METIS subgraphs for pretraining. As shown in Figure ~\ref{fig:scalilng}, GSPT's performance on both node classification and link prediction improves as more data is used for pretraining.  This demonstrates the potential for further enhancement of our pretraining method if industry-scale data becomes available (Answer to \textbf{RQ4}).

\section{Conclusion}

In this work, we present a novel graph pretraining framework based on random walks and a standard Transformer architecture. Building upon the unified feature space provided by LLM embeddings, we leverage masked feature reconstruction to perform fully self-supervised learning to learn transferrable node representations across different graphs. By pretraining on massive-scale graphs with over 100 million nodes, our framework demonstrates impressive transferability and achieves promising performance on both node-level and link-level tasks. Our findings enhance the comprehension of the LLM-unified feature space in graph data and offer valuable insights for the creation of a versatile and generalizable graph foundation model.

\bibliographystyle{plain}
\bibliography{ref}


\appendix
\newpage
\section{Appendix}

\subsection{Extending GSPT to Link Prediction}
\label{sec:extension-link-prediction}

Our proposed framework can be readily adapted to the link prediction task with minor adjustments. The key consideration is to construct an appropriate context for link prediction in a sequence-based manner to accomodate the Transformer architecture. Unlike node classification, which necessitates precise feature propagation, link prediction emphasizes modeling the neighborhood information of nodes such as the number of common neighbors and their distribution~\cite{mao2024revisiting}.
To achieve this, we adopt the 'Hop2Token'~\cite{Chen2022NAGphormerAT} approach for constructing the input context, i.e., \( h^0 = [x_0, x_1, \ldots, x_{l-1}] \), where \( x_0 \) represents the feature of the center node and \(x_k\) the aggregated feature of its K-hop neighborhood. This method ensures that the input sequence encapsulates the necessary neighborhood information for effective link prediction.

 To better capture the global connectivity in the graph, we incorporate a structure-aware decoder on top of the Transformer. This decoder takes the output of the Transformer and integrates them based on the graph's connectivity structure, enhancing the model's ability to predict links by leveraging global graph properties.
 Formally, the loss function for pretraining is computed by:
 \begin{equation}
    H^L = \text{TRM}(H^0)
 \end{equation}
\begin{equation}
    H_{node} = \text{Pooling}(H^L)
 \end{equation}
 \begin{equation}
    H^D = \text{Dec}(H_{node}, A)
 \end{equation}
 \begin{equation}
    \mathcal{L}_{\text{link}} = \frac{1}{m} \sum_{j=1}^{m} \left(1 - \cos(h^D_{i}, x_{i})\right)
 \end{equation}
 \text{Pooling} obtains the node representations by concatenating the output of the 0-hop (center) node and the averaged outputs of all other hops. \(\text{Dec}(H^L, A)\) indicates a structure-aware decoder that takes node representations and the adjacency matrix \(A\) as input, and outputs the decoded embeddings \(H^D\). For simplicity, we use \(h_{i}^D\) to denote the i-th row of \(H^D\). 
 The difference from node-level task is that (1) \(H^0\) is constructed with multi-hop aggregated features rather than independent nodes and (2) a decoder is used before computing the reconstruction loss. In practice, we adopt a one-layer GCN as the decoder.

\subsection{Related Works}
\textbf{Graph self-supervised learning.}
Limited by feature heterogeneity, studies in graph SSL are often limited to individual graphs ~\cite{hou2022graphmae, xu2021self, you2020graphGCL}, or focus on specific applications in biology and chemistry~\cite{hu2019strategies, xia2023molebert, Rong2020SelfSupervisedGT}, where a fixed-size 'vocabulary' allows for consistent representation of node or edge features across different graphs. Alternatively, some approaches entirely discard the original features to facilitate knowledge transfer based solely on structural information ~\cite{qiu2020gcc, xu2023better, davies2023its}, leading to sub-optimal performance. These constraints significantly hinder the broader application of graph SSL in real-world scenarios, particularly in developing a versatile and generalizable model.

\textbf{Feature-centric graph learning.}
To mitigate feature heterogeneity, ~\cite{oneforall} propose converting varied node features into text and training the model within this unified text space instead of the original vector spaces. Among the text-based methodologies, ~\cite{Chen2023ExploringTP, he2023harnessingTAPE} use frozen large language models (LLMs) to create fixed text embeddings for nodes across different graphs. In contrast, ~\cite{Chien2021NodeFE, Zhao2022LearningOL} implement a cascaded architecture that combines Language Models (LMs) and Graph Neural Networks (GNNs) to directly learn graph-aware text embeddings from original text attributes and graph topology. These approaches have demonstrated improved performance on various TAGs tasks, highlighting the critical importance of high-quality features in addition to graph structures.

\textbf{Graph foundation models.}
Developing a graph foundation model~\cite{mao2024graph} with a unified architecture that can adapt to diverse downstream tasks has been a popular research topic in the graph domain. 
Based on a LM-unified feature space,
OneForAll~\cite{oneforall} trains a model with a shared backbone across different data and tasks by transforming them into the same form, but its scope is limited to small-scale supervised learning. Prodigy~\cite{huang2023prodigy} focuses on empowering graph models with the ability to "learn in context" through pretraining, yet its specialized training strategy restricts it to few-shot inference and limited data utilization due to computational inefficiency. Meanwhile, these methods still rely on message passing, which has been shown to have limitations when generalizing to graphs with different structural or attribute properties~\cite{mao2024demystifying, lin2022spectral}.
In contrast to previous works, our proposed GSPT (1) presents a novel strategy to pretrain a backbone model across graphs without requiring supervision; (2) scales to massive scale graphs with efficiency and effectiveness; (3) showcases a message passing-free pretraining strategy, opening up new avenues for graph foundation model development.

\subsection{Datasets}
\label{sec:dataset}

\textbf{Summary.} The summary of datasets used in the experiments are presented in Table~~\ref{tab:datasets}. 

\begin{table}[ht]
\centering
\caption{Summary of datasets}
\begin{tabular}{ccc}
    \toprule
    \textbf{Name} & \textbf{\#Nodes} & \textbf{\#Edges} \\
    \midrule
    Cora & 2,708 & 10,858 \\
    Citeseer & 3,186 & 8,554 \\
    Pubmed & 19,717 & 88,670 \\
    Arxiv23 & 46,198 & 78,548 \\
    ogbn-papers100M & 111,059,956 & 1,615,685,872 \\
    \bottomrule
\end{tabular}
\label{tab:datasets}
\end{table}

\textbf{Partition.} For ogbn-papers100M, we use the \href{http://glaros.dtc.umn.edu/gkhome/views/metis}{METIS} algorithm to partition the graph into 11105 non-overlapping subgraphs. The statistics of the subgraphs are listed in Table ~\ref{table:partitions}. We use the implementation by \href{https://docs.dgl.ai/guide/distributed-partition.html}{dgl} in our experiments.

\begin{table}[ht]
    \centering
    \caption{Summary of METIS partitions on ogbn-papers100M}  
    \label{table:partitions}
    \begin{tabular}{c|c|c|c|c}
        \toprule
        \textbf{\#Graphs} & \textbf{Avg. \#Nodes} & \textbf{Avg. \#Edges} & \textbf{\#Node Range} & \textbf{\#Edge Range} \\
        \midrule
        11105 & 10000.90 & 61357.03 & 303 - 45748 & 328 - 122644 \\
        \bottomrule
    \end{tabular}
\end{table}

\textbf{Properties.} The graphs used in the experiments exhibit varying properties. To characterize these properties, we compute four commonly used graph metrics: average degree, sparsity, clustering coefficient, and homophily. The results are summarized in Table ~\ref{tab:dataset_property}. For the ogbn-papers100M dataset, we compute the average statistics using 100 randomly sampled METIS subgraphs. Note that ogbn-papers100M does not contain label data, so the homophily cannot be obtained.

\begin{table}[ht]
    \centering
    \caption{Summary of graph properties}
    \label{tab:dataset_property}
    \begin{tabular}{c|c|c|c|c}
        \toprule
        \textbf{Dataset} & \textbf{Average Degree} & \textbf{Homophily} & \textbf{Clustering Coefficient} & \textbf{Graph Density} \\
        \midrule
        Cora & 8.02 & 0.81 & 0.24 & 0.00144 \\
        Citeseer & 5.37 & 0.79 & 0.14 & 0.00083 \\
        Pubmed & 8.99 & 0.80 & 0.06 & 0.00023 \\
        Arxiv23 & 3.40 & 0.65 & 0.05 & 0.00004 \\
        Papers100M & 13.77 & N/A & 0.16 & 0.00173 \\
        \bottomrule
    \end{tabular}
\end{table}

\subsection{Accessibility}

\subsubsection{Dataset}

Citeseer: \url{https://github.com/CurryTang/Graph-LLM}

Cora, Pubmed: \url{https://github.com/kimiyoung/planetoid}

ogbn-papers100M: \url{https://ogb.stanford.edu/docs/nodeprop/}

Arxiv23: \url{https://github.com/XiaoxinHe/tape_arxiv_2023}

\subsubsection{Code}

\url{https://github.com/SongYYYY/GSPT}

\subsection{Experiment details}
\subsubsection{Hyperparameters}

\textbf{GSPT.} The hyperparameters used for GSPT are listed in Table ~\ref{tab:hyper-node}, Table ~\ref{tab:hyper-link}, and Table ~\ref{tab:hyper-finetune}. GSPT-node indicates the version for node classification, while GSPT-link denotes the one for link prediction. GSPT-link also involves fine-tuning the pretrained model on individual datasets. 

\begin{table}[ht]
    \centering
    \caption{Hyperparameters of GSPT-node}
    \label{tab:hyper-node}
    \resizebox{\linewidth}{!}{
    \begin{tabular}{  l | l | l  }
        \toprule
        \textbf{Hyperparameter} & \textbf{Value} & \textbf{Explanation} \\
        \midrule
        mask\_rate & 0.2 & Probability of masking a node \\
        p\_random & 0.2 & Probability of replacing [MASK] with random nodes, see~\cite{Devlin2019BERTPO} \\
        p\_unchanged & 0.2 & Probability of keeping masked nodes unchanged, see~\cite{Devlin2019BERTPO} \\
        hidden\_dim & 768 & Dimension of hidden layers \\
        ffn\_dim & 3072 & Dimension of feed-forward network layers \\
        n\_layers & 3 & Number of Transformer layers \\
        n\_heads & 12 & Number of attention heads \\
        epochs & 10 & Number of training epochs \\
        weight\_decay & 0.01 & strength of L2 regularization \\
        peak\_lr & 0.0001 & Peak learning rate \\
        end\_lr & 0.00001 & End learning rate (after decay) \\
        warmup\_updates
        & 10000 & Number of warmup updates \\
        dropout & 0.3 & Dropout rate \\
        attention\_dropout & 0.3 & Dropout rate for attention layers \\
        emb\_dropout & 0.3 & Dropout rate for embeddings \\
        p & 0.25 & Return parameter of random walk, see ~\cite{Grover2016node2vecSF} \\
        q & 0.25 & In-out parameter of random walk, see ~\cite{Grover2016node2vecSF} \\
        walk\_length & 20 & Length of random walk \\
        \bottomrule
    \end{tabular}}
\end{table}

\begin{table}[ht]
    \centering
    \caption{Hyperparameters of GSPT-link}
    \label{tab:hyper-link}
    \resizebox{\linewidth}{!}{
    \begin{tabular}{  l | l | l  }
        \toprule
        \textbf{Hyperparameter} & \textbf{Value} & \textbf{Explanation} \\
        \midrule
        mask\_rate & 0.5 & Probability of masking a node \\
        p\_random & 0 & Probability of replacing [MASK] with random nodes, see~\cite{Devlin2019BERTPO} \\
        p\_unchanged & 0 & Probability of keeping masked nodes unchanged, see~\cite{Devlin2019BERTPO} \\
        hidden\_dim & 384 & Dimension of hidden layers \\
        ffn\_dim & 768 & Dimension of feed-forward network layers \\
        n\_layers & 2 & Number of Transformer layers \\
        n\_heads & 8 & Number of attention heads \\
        epochs & 100 & Number of training epochs \\
        weight\_decay & 0 & strength of L2 regularization \\
        peak\_lr & 0.001 & Peak learning rate \\
        end\_lr & 0.0001 & End learning rate (after decay) \\
        warmup\_updates
        & 100 & Number of warmup updates \\
        dropout & 0.1 & Dropout rate \\
        attention\_dropout & 0.1 & Dropout rate for attention layers \\
        emb\_dropout & 0 & Dropout rate for embeddings \\
        n\_hops & 3 & Number of hops for feature aggregation \\
        \bottomrule
    \end{tabular}}
\end{table}

\begin{table}[ht]
    \centering
    \caption{Hyperparameters of GSPT-link, fine-tuning.}
    \label{tab:hyper-finetune}
    \resizebox{\linewidth}{!}{
    \begin{tabular}{  l | l | l }
        \toprule
        \textbf{Hyperparameter} & \textbf{Value} & \textbf{Explanation} \\
        \midrule
        lr & [1e-3, 1e-4] & Learning rate  \\
        projector\_layers & 3 & Number of layers in the projector module to compute edge scores \\
        projector\_dim & 256 & Dimension of the projector module to compute edge scores \\
        epochs & 1000 & Number of training epochs \\
        patience & 20 & Patience of early stopping \\
        batch\_size & 4096 & Number of samples (edges) per batch \\
        \bottomrule
    \end{tabular}}
\end{table}

\textbf{Baselines.} For all baselines, we search the optimal hyperparameters using the validation set of individual datasets. We organize the baselines based on the method and task type into the following.

\textbf{GNNs for few-shot node classification.} 
Table ~\ref{tab:hyper-gnn} contains the details for reproducing the results for end-to-end methods in Table ~\ref{tab:node_performance_comparison}. Such methods involve training specific models on individual downstream datasets.

\begin{table}[ht]
    \centering
    \caption{Hyperparameters of GNNs on few-shot node classification}
    \label{tab:hyper-gnn}
    \resizebox{\linewidth}{!}{
    \begin{tabular}{ l | l | l | l | l | l | l }
        \toprule
        \textbf{Hyperparameter} & lr & weight\_decay & hidden\_dim & dropout & num\_layers & num\_heads \\
        \midrule
        \textbf{Search Range} & [1e-2, 1e-3] & [0, 1e-4, 5e-4] & [64, 256, 384] & [0, 0.5] & [2] & [4, 8] \\
        \bottomrule
    \end{tabular}}
\end{table}

\textbf{SSLs for few-shot transfer.} Table~\ref{tab:hyper-ssl} contains details of pretraining with graph self-supervised learning methods on ogbn-papers100M. Like in GSPT, we perform pretraining on the METIS partitions of the entire graph. These pretrained backbones are then used for generating node representations on the downstream datasets to produce the results in Table~\ref{tab:node_performance_comparison}.

\begin{table}[ht!]
\centering
\caption{Hyperparameters for self-supervised transfer learning methods.}
\label{tab:hyper-ssl}
\begin{tabular}{l|l|l} 
 \toprule
 \textbf{Hyperparameter} & \textbf{DGI} & \textbf{GraphMAE} \\ 
 \midrule
 Backbone & GCN & GAT \\ 
 num\_layers & 1 & 2 \\
 num\_heads & - & 4 \\ 
 lr & 0.001 & 0.001 \\
 weight\_decay & 0 & 0 \\
 dropout & 0 & - \\
 feat\_drop & - & 0.2 \\
 attn\_drop & - & 0.1 \\
 hidden\_dim & 384 & 384 \\
 nonlinearity & prelu & prelu \\
 epochs & 10 & 10 \\
 mask\_rate & - & 0.5 \\
 alpha & - & 3 \\
 \bottomrule
\end{tabular}
\end{table}

\textbf{GNNs for link prediction.} Table~\ref{tab:hyper-link-gnn} presents the hyperparameters of GNN baselines for link prediction to reproduce results in Table~\ref{tab:link_performance_comparison}. Specifically, we adopt an MLP on top of the GNNs that takes the product of node representations as input, and output the score of edge existence.

\begin{table}[ht!]
\centering
\caption{Hyperparameters for link prediction baselines.}
\label{tab:hyper-link-gnn}
\begin{tabular}{l|l|l} 
 \toprule
 \textbf{Model} & \textbf{Hyperparameter} & \textbf{Search Range} \\ 
 \midrule
 \multirow{3}{*}{GNN} & num\_layers & [1, 2, 3] \\ 
 & hidden\_dim & [128, 256] \\ 
 & num\_heads & [1, 4] \\ 
 \midrule
 \multirow{2}{*}{MLP} & num\_layers & [1, 2, 3] \\ 
 & hidden\_dim & [128, 256] \\ 
 \midrule
 \multirow{5}{*}{General} & lr & [1e-3, 1e-2] \\ 
 & weight\_decay & [0, 1e-5] \\ 
 & dropout & [0.1, 0.5] \\
 & batch\_size & 4096 \\
 & epochs & 1000 \\ 
 & patience & 20 \\ 
 \bottomrule
\end{tabular}
\end{table}


\end{document}